\documentclass{optica-article}

\journal{opticajournal}

\articletype{Research Article}

\usepackage{lineno}
\usepackage{cite}
\usepackage{color}
\usepackage[ruled,linesnumbered]{algorithm2e}
\usepackage{caption}
\usepackage{booktabs} 
\usepackage{cleveref}
\usepackage{graphicx}
\usepackage{listings}
\usepackage{hyperref}
\usepackage{epstopdf}
    
\usepackage{multirow,makecell}
\usepackage{bm}
\usepackage{subcaption}

\DeclareUnicodeCharacter{2212}{-}
\usepackage{dsfont}

\hypersetup{
    colorlinks=true,
    linkcolor=blue,
    citecolor=blue,
    urlcolor=blue
}

\begin{document}

\title{PiLocNet: Physics-informed neural network on 3D localization with rotating point spread function}

\author{
Mingda Lu,\authormark{1} 
Zitian Ao,\authormark{2} 
Chao Wang,\authormark{2,3,*} 
Sudhakar Prasad,\authormark{4} and 
Raymond Chan\authormark{5,*}
}

\address{
\authormark{1}Department of Mathematics, City University of Hong Kong, Hong Kong SAR, China\\
\authormark{2}Department of Statistics and Data Science, Southern University of Science and Technology, Shenzhen 518005, Guangdong Province, China\\
\authormark{3}National Centre for Applied Mathematics Shenzhen, Shenzhen 518055, Guangdong Province, China\\
\authormark{4}School of Physics and Astronomy, University of Minnesota, USA\\
\authormark{5}Lingnan University, Hong Kong SAR, China
}

\email{
\authormark{*}wangc6@sustech.edu.cn; raymond.chan@ln.edu.hk\\
} 

\begin{abstract}
For the 3D localization problem using point spread function (PSF) engineering, we propose a novel enhancement of our previously introduced localization neural network, LocNet. The improved network is a physics-informed neural network (PINN) that we call PiLocNet. Previous works on the localization problem may be categorized separately into model-based optimization and neural network approaches. Our PiLocNet combines the unique strengths of both approaches by incorporating forward-model-based information into the network via a data-fitting loss term that constrains the neural network to yield results that are physically sensible. We additionally incorporate certain regularization terms from the variational method, which further improves the robustness of the network in the presence of image noise, as we show for the Poisson and Gaussian noise models. This framework accords interpretability to the neural network, and the results we obtain show its superiority. Although the paper focuses on the use of a single-lobe rotating PSF to encode the full 3D source location, we expect the method to be widely applicable to other PSFs and imaging problems that are constrained by well modeled forward processes.
\end{abstract}




\section{Introduction} \label{introduction}

\begin{figure}[htbp]
    \centering
        \includegraphics[width=1\textwidth]{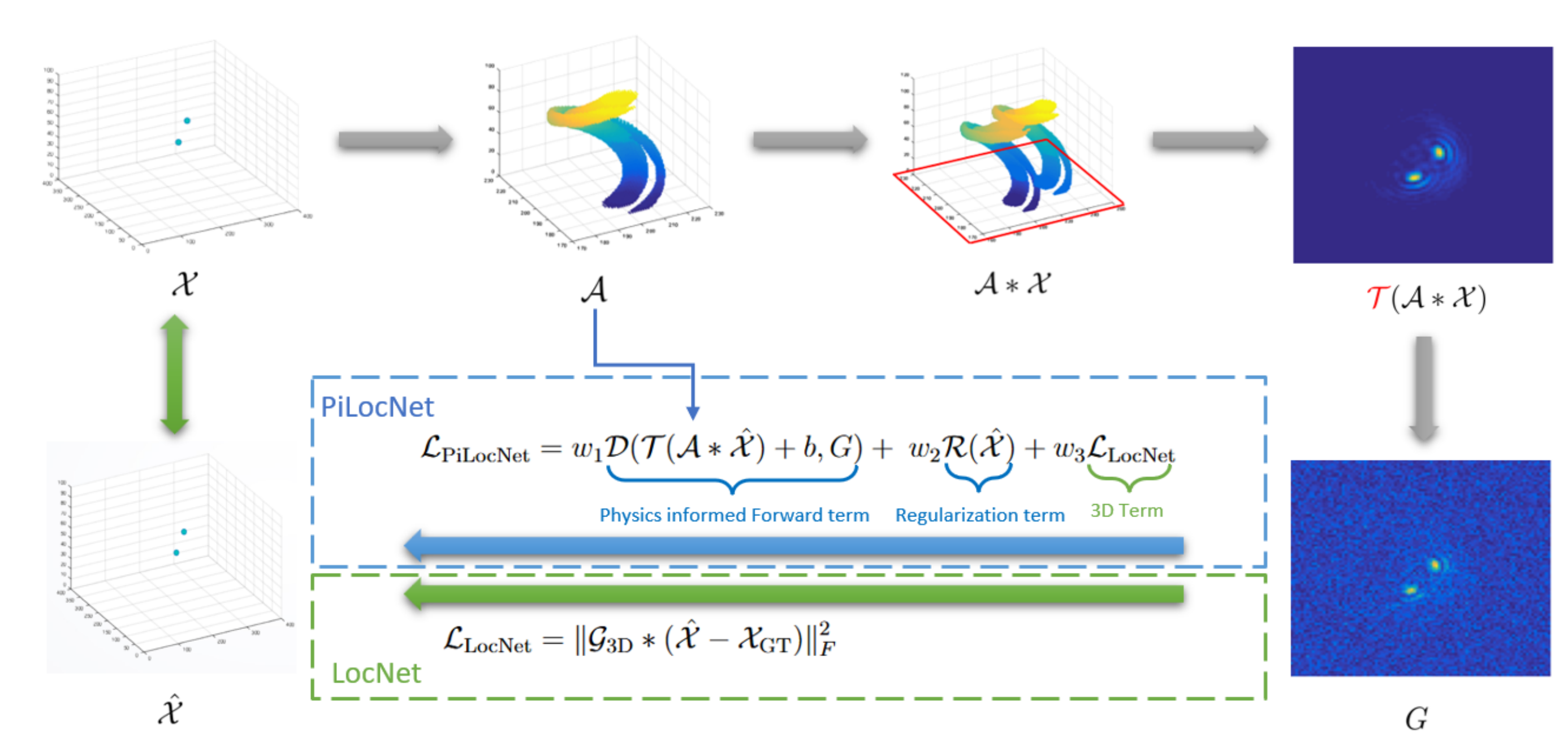}
        \caption{Overview of the PiLocNet: the main improvement is in the inclusion of a forward loss term based on the matrix $\mathcal{A}$ that models the 3D point spread function along with an appropriate regularization term into the loss function. This added physics information improves network training.}
        \label{PiLocNet_Overview}
    \medskip
\end{figure}

Locating point sources or structures in three-dimensional (3D) space is a common challenge in many scientific applications. This is particularly relevant in computer vision, which has numerous applications like robotics, augmented reality, and autonomous systems. For solving the 3D localization problem, point spread function (PSF) engineering is a promising and effective technique that places a specific phase function into the imaging aperture.  This aperture function processes the photons emitted by a source point and entering the imaging system into an image pattern that carries information about the full 3D location of the source. Unlike traditional methods that leverage stereo or multi-view images,  PSF engineering imprints the 3D location information of the point sources into the position and form of the corresponding images acquired on a single two-dimensional (2D) sensor. 
PSF-associated methodologies have wide applications that range from telescopic to microscopic imaging systems and have significantly enhanced the precision of point source localization. For example, single-molecule localization microscopy (SMLM) \cite{RN57} localizes individual fluorophores in 3D structures to render super-resolution imaging of fluorescent molecules. By leveraging the $z$-dependent form of the PSFs, 3D SMLM surpasses traditional diffraction limits, allowing for the visualization of biological structures at near-molecular resolution in all three dimensions. Numerous field studies have underscored this methodology's significance and potential \cite{ marsh2018artifact, Min2014FALCONFA, Boyd2018DeepLocoF3, Kikuchi2022PhotochemicalMO}.

Various depth-encoding phase masks have been developed for PSF engineering, yielding a variety of PSFs, such as astigmatic \cite{RN31}, double helix (DH) \cite{RN32}, and tetrapod \cite{RN30}. Here, we mainly consider the single-lobe rotating PSF (RPSF) invented by Prasad \cite{RN29}. By means of a suitable spiral phase function in the imaging aperture, one can create a PSF that rotates by an angle proportional to the depth ($z$) coordinate of the point source around a center fixed by the 2D transverse $(x,y)$  coordinates of the source. The rotating PSF comprises a single bright lobe surrounded by a fainter ring-shaped substructure, which rotates together as the source defocus distance along the optical ($z$) axis changes. One of the main benefits of a single-lobe rotating PSF over other more complicated PSFs like the DH and tetrapod PSFs is that the former concentrates the photon energy into its single lobe with a higher flux density, making it more noise-robust in crowded source fields \cite{RN33}.

The main objective of the problem is to recover the 3D locations of point sources from their observed noisy image as accurately as possible.  Various methods have been proposed and categorized into mathematical optimization and neural network approaches. 
 Several variational methods \cite{RN59, RN48, RN47, RN34} have been recently introduced from a non-convex optimization perspective. In the case of Gaussian image noise, the Frobenius norm is used as the data fitting term, and a continuous exact $\ell_0$ penalty (CEL0) is the regularization term.  For the case of Poisson noise, the KL-NC model \cite{RN34} uses the Kullback-Leibler (KL) divergence data-fitting term and a different non-convex regularization term. The optimization problem is solved by an iteratively reweighted  $ \mathrm{\ell_1} $ algorithm. 

Deep-learning neural network approaches have also been proposed to solve this 3D localization problem. Two important architectures are DeepSTORM3D \cite{RN26} and DECODE \cite{RN51}. Specifically, DeepSTORM3D uses a 3D grid network to map and predict the coordinates of the point sources, while DECODE uses a different structure, with multiple channels, to predict different kinds of information about the input images, including the 3D coordinates, brightness, and the probability of existence of point sources. Recently, Dai et al. proposed LocNet \cite{RN62}, adapting DeepSTORM3D on single-lobe rotating-PSF images, with an additional post-processing step to cluster the initial prediction of the network. 

In the field of neural network studies, the method of physics-informed neural network (PINN) has emerged in recent years \cite{RN63,PINN_micro,wang2024deep,tsakyridis2024photonic}, initially in the context of problems involving partial differential equations (PDEs) but later applied more widely. The goal is to incorporate any known physics information about the problem into the neural network structure or loss function. 

Inspired by the idea of PINN, we propose here the Physics-Informed Localization Network (PiLocNet) method for the PSF localization problem, as shown schematically in Fig.~\ref{PiLocNet_Overview}. Given the process of PSF image generation, the forward model is a known piece of physical information.  Supplying such physical information to the neural network is helpful to the network in generating better results than a random black-box kind of fitting approach characteristic of more conventional neural networks. 
With model-specific data fitting and regularization terms, a PINN-based method is interpretable to the neural network. 
The numerical experiments we report here have been conducted based on the RPSF model. Our results, as we will show, prove that the added physical information can significantly improve the prediction accuracy in terms of both precision and recall rates. Our ablation studies also verified the robustness of PiLocNet. 

The rest of this paper is organized as follows: In Section~\ref{sec:single_lobe}, we briefly review the optical model of the rotating PSF and the variational methods we used previously for this localization problem. In Section~\ref{sec:method}, we introduce the specific model structure, including an improved loss function that encompasses both the forward model and regularization terms. Next, we introduce a series of simulation-based experiments that we conducted to verify the effects of our PiLocNet in Section~\ref {sec:results}.  We conclude the paper with a summary of our findings and future work in Section~\ref{sec:conclusion}.

\section{Single-lobe point spread function and its noise models}\label{sec:single_lobe}

This section will provide a review of the single-lobe rotating PSF forward model and the two different noise models that we explore in the paper.

\subsection{Forward model of single-lobe RPSF} \label{forward_model}

\begin{figure}[htbp]
    \centering
        \includegraphics[width=1\textwidth]{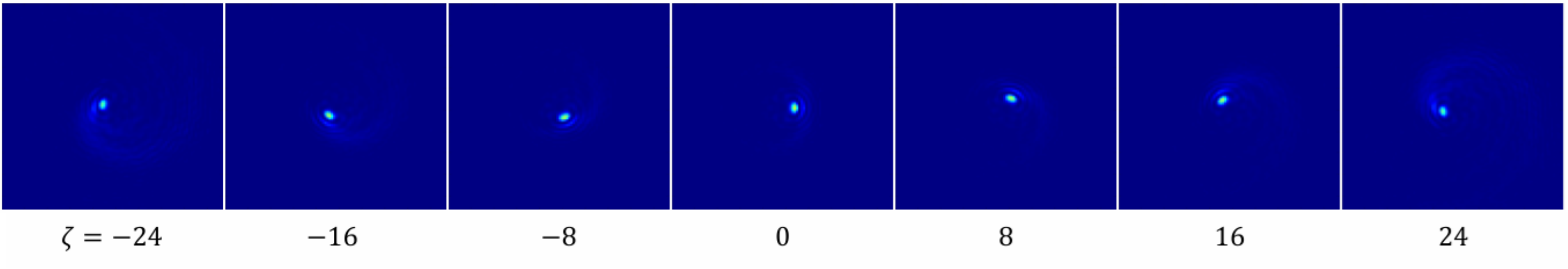}
        \caption{single-lobe point spread function applied images with different depth $\zeta$.}
        \label{RPSF_depth}
    \medskip
\end{figure}

The forward model has been described in great detail in our previous papers \cite{RN29,RN47,RN34}. 
Here, we only present a brief summary of the model. The RPSF image, $\mathcal{A}_\zeta$, for a point source with defocus parameter $\zeta$, a unit flux $f=1$, at the source location $\mathbf{r}_O=(x_O,y_O)$ is given by:
\begin{equation*}
    A_\zeta(\mathbf{s})=\frac{1}{\pi}\left|\int_{\Omega}\mathrm{exp}\left[i\left(2\pi\mathbf{u}\cdot\mathbf{s}+\zeta u^2-\psi(\mathbf{u})\right)\right]d\mathbf{u}\right|^2,
\end{equation*}
where $\zeta=-\frac{\pi\delta z R^2}{\lambda z_O(z_O+\delta z)}$, $\Omega$ represents the circular disk-shaped clear pupil and $i=\sqrt{-1}$. The quantity, $\mathbf{s}=\frac{\mathbf{r}}{\lambda z_I/R}$, is the position vector, $\mathbf{r}$, of an image-plane point relative to the Gaussian image location, when expressed in units of the Rayleigh diffraction scale, $\lambda z_I/R$, in which 
$\lambda$ is the imaging wavelength,  $R$ is the radius of the pupil, and $\delta z$, $z_O$, $z_I$ are the distances from the object plane to the in-focus object plane, the in-focus object plane to the pupil plane, and the pupil plane to the image plane, respectively. 
The symbol $\mathbf{u}$ denotes the position vector in the plane of the pupil, in units of the radius of the pupil. Its polar coordinates are $\mathbf{u}=(u,\phi_u)$. The circular pupil is segmented into $L$ different contiguous annular Fresnel zones, with each zone carrying a spiral phase function, $\psi(\mathbf{u})$, with the number of complete phase cycles changing successively by 1 from one zone to the next.  
The RPSF can be shown \cite{RN29} to continuously rotate within the scaled defocus range, $\zeta\in[-\pi L,\pi L]$, as it begins to spread out, break apart, and lose its shape unacceptably outside this range. An illustration of the images of a single point source at different values of the depth parameter $\zeta$, when the RPSF is used, is presented in Fig ~\ref{RPSF_depth}.

With the above formulation, for $N$ point sources the observed image data count $G(x,y)$ at location $(x,y)$  may be expressed as:
\begin{equation*}
    G(x,y)\approx\mathcal{N}\left(\sum_{i=1}^{N}{A_i(x-x_i,y-y_i)f_i}+b\right),
\end{equation*}
where $(x_i,y_i)$ and $f_i$ are the transverse coordinates and flux of the $i$th point source. Its depth coordinate, $z_i$, is embedded, via the depth-parameter value $\zeta_i$, in the PSF $A_i$ 
and $\mathcal{N}$ is the operator for incorporating noise. 

Specifically, the forward model for the Gaussian noise case can be conceptualized as following the Gaussian distribution at the $p$-th pixel, 

\begin{equation}\label{forward model Gaussian}
    G_p \sim \mathbb{N}( [\mathcal{T}(\mathcal{A}*\mathcal{X})]_p + b, \sigma^2), \quad p = 1, 2, ..., d,
\end{equation}
where $\mathbb{N}(\mu, \sigma^2)$ denotes the Gaussian distribution with expectation $\mu$ and variance $\sigma^2$, $\mathcal{A}*\mathcal{X}$ is the 3D convolution of $\mathcal{A}$ with $\mathcal{X}$, and $\mathcal{T}$ projects out a 2D slice of the convolution. $b$ represents the background photons, which is always a constant no matter how $\sigma^2$ changes in Gaussian noise cases. The symbol $\mathcal{A}$ denotes the 3D PSF dictionary represented as a cube, which is built from a series of images, each corresponding to a different depth, while $\mathcal{X}$  
contains the 3D coordinates of the point sources where each entry's value is the corresponding source flux. 
The total number of pixels in the 2D image array is $d=H\times W$.
The forward model for the case of Poisson noise closely resembles that for Gaussian noise, differing solely in the method of generation of noise:
\begin{equation}\label{forward model Poisson}
    G_p \sim \mathbb{P}([\mathcal{T}(\mathcal{A}*\mathcal{X})]_p + b), \quad p = 1, 2, ..., d,
\end{equation}
where $\mathbb{P}(\lambda)$ denotes the Poisson distribution with expectation $\lambda$. 


\subsection{Optimization approach: the variational models  } \label{variational models}



In order to recover the 3D tensor $\mathcal{X}$ from the given observed image $G$, the optimization approach can be formulated as a minimization problem,
\begin{equation*}
\min_{\mathcal{X}}{\mathcal{D}}\left(\mathcal{T}\left(\mathcal{A}\ast\mathcal{X}\right)+b,G\right)+\mathcal{R}\left(\mathcal{X}\right),
\end{equation*}
where $\mathcal{D}$ enforces data fitting and $\mathcal{R}\left(\mathcal{X}\right)$ is an appropriate regularization term. We next formulate these two terms for our two different noise models. 

\subsubsection{The case of Gaussian noise}\label{sec:Gaussian}

Gaussian noise is a common type of noise for which the random error, as we have just noted, has a Gaussian probability distribution. This noise is present due to various factors such as sensor read-out error, non-uniform brightness response of the image sensor, noise and mutual interference from circuit components, and prolonged usage of the image sensor at high temperatures. 
For the Gaussian noise case, the data-fitting term is a simple quadratic function, 
\begin{equation*}
    \mathcal{D}(\mathcal{T}(\mathcal{A} * \mathcal{X}) + b, G) :=  \left\| \mathcal{T}(\mathcal{A} * \mathcal{X}) + b - G \right\|_F^2,
\end{equation*}
where $\|\cdot \|_F$ denotes the Frobenius norm, namely the $\ell_2$ norm, of the vectorized input. The regularization term enforces sparsity, for which we use a non-convex term approaching the $\ell_0$ norm for linear least squares data fitting problems. Specifically, we have used the Continuous Exact $\ell_0$ (CEL0) penalty function \cite{CEL0} defined as: 
\begin{equation} \label{eq:cel0}
    \mathcal{R}(\mathcal{X}) := \phi_{\rm CEL0}(\mathcal{X}) = \sum_{i,j,k=1} \phi(\|\mathcal{T}(\mathcal{A} * \delta_{ijk})\|, \mu; \mathcal{X}_{ijk}),
\end{equation}
where $\phi(a, \mu; u) = \mu - \frac{a^2}{2} \left( |u| - \frac{\sqrt{2\mu}}{a} \right)^2 \mathds{1}_{\{|u| \leq \frac{\sqrt{2\mu}}{a}\}}$ and $\mathds{1}_{E} := \begin{cases} 
1 & \text{if } u \in E; \\ 
0 & \text{others}.
\end{cases}$. In addition, $\delta_{ijk}$ is a 3D tensor whose only nonzero entry is at $(i,j,k)$ with value 1; $\mu$ is the parameter to control the non-convexity. 
The minimization problem was formulated as
\begin{equation*}
\min_{\mathcal{X} \geq 0} \left\{  \|\mathcal{T}(\mathcal{A} * \mathcal{X}) + b - G\|_F^2 + \sum_{i,j,k=1} \phi(\|\mathcal{T}(\mathcal{A} * \delta_{ijk})\|, \mu; \mathcal{X}_{ijk}) \right\}.
\end{equation*}

\subsubsection{The case of Poisson noise}\label{sec:Poisson}
The Poisson noise model describes the probability distribution of the number of random events, such as photon counts, occurring per unit time. The data fitting term for the Poisson noise case is the \( I \)-divergence, which is also known as the Kullback-Leibler (KL) divergence \cite{klnc1}:
\begin{equation*}
    \mathcal{D}(\mathcal{T}(\mathcal{A} * \mathcal{X}) + b, g) := D_{KL}(\mathcal{T}(\mathcal{A} * \mathcal{X}) + b, G),
\end{equation*}
where \( D_{KL}(z, g) = \langle g, \ln \frac{g}{z} \rangle + \langle 1, z - g \rangle \). The sparsity-enforcing regularization term is designed as a non-convex function \cite{klnc2,klnc3,klnc4}:
\begin{equation*}
    \mathcal{R}(\mathcal{X}) := \mu \sum_{i,j,k=1} \theta(a; \mathcal{X}_{ijk}) = \mu \sum_{i,j,k=1} \frac{|\mathcal{X}_{ijk}|}{a + |\mathcal{X}_{ijk}|},
\end{equation*}
where \( a \) is a fixed parameter that determines the degree of non-convexity. The Poisson minimization problem was formulated as
\begin{equation*} \label{eq:Poisson_model}
    \min_{\mathcal{X} \geq 0} \left\{ (1, \mathcal{T}(\mathcal{A} * \mathcal{X}) - G \ln(\mathcal{T}(\mathcal{A} * \mathcal{X}) + b)) + \mu \sum_{i,j,k=1} \frac{|\mathcal{X}_{ijk}|}{a + |\mathcal{X}_{ijk}|} \right\}.
\end{equation*}

\section{The PINN Methodology}\label{sec:method}

Here we propose a physics-informed neural network called PiLocNet that works for RPSF imaging for the Gaussian and Poisson noise models. As an enhancement of the typical black-box type of neural networks, the proposed model builds the known physics information of the forward process into a PINN framework through additional loss functions. 


 

\subsection{Convolutional Neural Network: LocNet}

LocNet\cite{RN26}, which combines a deep convolutional neural network (CNN) with a post-processing step, was adopted for RPSF-image based 3D source localization. The CNN part, similar to DeepSTORM3D, consists of 3D grid layers to accommodate the point source prediction. Several practical CNN techniques were employed, such as up-sampling and residual layers. The loss function of LocNet is
\begin{equation*}
\mathcal{L}_{\rm LocNet}=\|{\cal G}_{\rm{3D}}\ast (\hat{\cal X}-{\cal X}_{\rm GT} )\|^2_F,
\end{equation*}
which is the mean square error of the ground truth ${\cal X}_{\rm GT}$ and prediction $\hat{\mathcal{X}}$, with both smoothed by a 3D Gaussian kernel ${\cal G}_{\rm{3D}}$. After the network generates the initial predictions, a post-processing step further refines the results by clustering closely spaced point sources into one and removing sources with brightness lower than a threshold.

\subsection{The pipeline of PiLocNet and its architecture} \label{LocNet v2}

LocNet is a data-driven approach that only considered the Poisson noise scenario in Ref. \cite{RN26}. Here we 
incorporate PINN into LocNet and propose a new framework, PiLocNet, for both Poisson and Gaussian noise cases. 
The main idea of PINN is to incorporate information about the physics of the problem at hand into the neural network's loss function, as illustrated in the Fig.~\ref{PiLocNet_Overview}. This approach helps the training process to achieve accurate results by minimizing the loss with improved guidance provided by known physical information. In this context, we discuss how this concept can be implemented to solve the PSF problem. We modify the LocNet loss function by adding to it two extra terms,
\begin{equation}
\label{eq:main_loss}
\mathcal{L}_{\rm PiLocNet}=w_1 \mathcal{D}(\mathcal{T}(\mathcal{A} * \hat{\mathcal{X}}) + b, G)+\ w_2 \mathcal{R}(\hat{\mathcal{X}})+w_3 \mathcal{L}_{\rm LocNet},
\end{equation}
the first term being the data-fitting term, which contains the PSF operator, $\mathcal{A}$, the known physics information to guide the neural network. However, to add the data fitting term into the loss function correctly, it needs to be model-dependent for different noise types and must be accompanied by an appropriate regularization, which is its second term. 
Additionally, we need different relative weights, $w_1: w_2:w_3$, for the three terms to ensure proper balance and trade-off of these terms. 


We employ the same data fitting and regularization terms for the cases of Gaussian and Poisson noise that we described in the previous section but now allow them to have different relative weights. In other words, we use the following PINN loss functions for the two noise cases, respectively:  
\begin{equation}\label{loss Gaussian}
\mathcal{L}_{g}=w_1\|\mathcal{T}(\mathcal{A}*{\hat{\mathcal{X}}})+ b-G\|_F^2+w_2\Phi_{\rm CEL0}(\hat{\mathcal{X}})+w_3 \|{\cal G}_{\rm{3D}}\ast (\hat{\cal X}-{\cal X}_{\rm GT} )\|^2_F. 
\end{equation}
and

\begin{equation}\label{loss Poisson}
\mathcal{L}_{p}=w_1\left\langle 1,  \mathcal{T}(\mathcal{A} \ast \hat{\mathcal{X}})- G \ln(\mathcal{T}(\mathcal{A} \ast \hat{\mathcal{X}})+b) \right\rangle+w_2\sum_{ijk}\tfrac{\lvert\hat{\mathcal{X}}_{ijk}\lvert}{\lvert\hat{\mathcal{X}}_{ijk}\lvert+a}+w_3 \|{\cal G}_{\rm{3D}}\ast (\hat{\cal X}-{\cal X}_{\rm GT} )\|^2_F. 
\end{equation}
For most of the experiments conducted in the paper, the weights $w_1, w_2, w_3$, were in the ratio 1:700:1000 for the case (\ref{loss Gaussian}) of Gaussian noise and 1:1:500 for the case (\ref{loss Poisson}) of Poisson noise. For experiments in ablation study on different noise levels, one can go through a searching process to reach optimal weight values correspondingly. 

The architecture of PiLocNet closely resembles that of LocNet \cite{RN62}, which was shown to be robust. In a hypothetical experiment with no noise added, the network can output results with very high accuracy, proving that it is well-designed. 
The network leverages convolutional kernels of specific dimensions to extract pertinent features, followed by batch normalization to expedite convergence. Subsequently, the ReLU activation function is applied. Finally, a shortcut connection is utilized to merge the input content with the output layer, facilitating residual convolution via a summation layer.
The final prediction layer consists of a convolution layer with a convolution kernel size of $\pi$ and an activation layer with an activation function with a HardTanh range of $\pi$. The final output is a 3D lattice image, where the value of each vertex reflects the degree of confidence that there is a point source near the point. The higher the value, the more likely a point source is near the lattice point.

From the output of the network, we obtain a tensor $ \hat{\mathcal{X}} \in \mathbb{R}^{H \times W \times D} $ as the initial prediction, where  $ H $ is the pixel size of height, $ W $ is the pixel size of width, and $ D $ is the pixel size of depth. 
The value of the tensor at a node is proportional to the probability of the existence of a real source near this node; we name the value of each node as intensity, denoted as $\hat{\mathcal{X}}_{ijk} \in [0, \pi]$. After initial prediction, we employ the same post-processing as in \cite{RN47} to $\hat{\mathcal{X}}$, which refines the results by clustering nearby points of 2 pixels, and removing points with brightness lower than a threshold set at 5\% of the highest value of the tensor. Then we obtain a set of points $\hat{\mathbf{X}}= \{ \mathrm{\hat{x}}_1, ..., \mathrm{\hat{x}}_m \}$, 
where $\mathrm{\hat{x}}_i \in \left( (0, H) \times (0, W) \times (0, D)\right) \subseteq  \mathbb{R}^3$ is a 3D location vector for each $i \in \{1, 2, ..., m\}$.  


\subsection{Network training}

The construction of our training dataset proceeds as follows. We generate a corpus comprising 10,000 images, each defined on a $96 \times 96$ pixel array. The flux values for the point sources within each image are drawn from a Poisson distribution with a mean of 2000 photon counts. 
Subsequently, $90\%$ of the images in this dataset are allocated for training purposes, while the remaining $10\%$ are used for validation. Within this cohort of 10,000 images, the number of point sources is randomly distributed following a uniform distribution from 5 to 50. For our test dataset, we introduce varying numbers, also referred to as densities, of point sources, specifically 5, 10, 15, 20, 25, 30, 35, 40, and 45 sources per image, with the aim to assess our model's efficacy across different source densities. Finally, we generate 100 images for each of these densities, thus 900 images in all, for comprehensive testing. We employ the Adam optimization algorithm \cite{kingma2014adam} in conjunction with a mini-batch size 16. The initial learning rate is stipulated at $1 \times 10^{-3}$, with a decay factor of 0.5 applied after every three epochs if there is no discernible improvement in the loss. The termination criterion for the training process is either the absence of any improvement in validation loss over 15 epochs or a validation loss lower than $1 \times 10^{-7}$. 
The training regimen spanned 200 epochs and utilized an NVIDIA A100-SXM4-40GB GPU.

\section{Results}\label{sec:results}

To evaluate the 3D-localization performance of our proposed method, we employ recall and precision rates. 
The \textbf{recall rate} is defined as the ratio of the total number of predicted true positives to the total number of point sources that should have been identified as positive. The \textbf{precision rate} is similarly defined as the ratio of the total number of true positives to the total number of point sources predicted as positive. True positives are identified based on a specified distance threshold between predicted and ground-truth point sources based on \cite{RN62}. Note that reducing false negatives improves recall, while reducing false positives improves precision.


\subsection{Comparison with previous methods} \label{PiLocNet to LocNet to LocNet v2}

Based on the experimental setup outlined previously, we assess the average recall and precision rates across three different methodologies: the variational methods \cite{RN47}, the original LocNet method \cite{RN62}, and a modified  LocNet v2, which has the same loss function as LocNet except that its architecture is changed to be that of PiLocNet. The primary changes we have made were the removal of the up-sampling layer and adjustments made to the dilation rates within the residual convolution layers. 
The decision to eliminate the up-sampling layer stemmed from our observation that its removal reduces training time substantially without compromising the model's performance. The dilation rates were aligned from $\{1, 2, 5, 9, 17\}$ to $\{1, 2, 4, 8, 16\}$.  
A comparative analysis between PiLocNet and LocNet v2 is crucial to ascertain the efficacy of our proposed model in terms of PiLocNet's use of a more physically sensible loss function. 
The chosen noise model is either Gaussian or Poisson, as described by Eq.~\ref{forward model Gaussian} or Eq.~\ref{forward model Poisson}, respectively. For the case of Gaussian noise, its standard deviation, $\sigma$, is taken to be uniform across the image and equal to a fraction of the value, $I_{\rm max}$, of the maximum flux at the pixels in an arbitrary observed image. Unless noted otherwise, we chose $\sigma = 0.1 \times I_{\rm max}$ for our studies on the Gaussian noise model and the background value of $b=5$ for both noise models. 

\begin{table}[ht]
\centering
\captionsetup{font={footnotesize,bf, sf,stretch=1},justification=centering}
\caption{Evaluation results of $\ell_2-\rm{CEL0}$, LocNet, LocNet v2, and PiLocNet for RPSF images with Gaussian noise }
\resizebox{\textwidth}{!}{
\begin{tabular}{c|cc|cc|cc|cc}
\hline
 & \multicolumn{2}{c|}{$\ell_2-{\rm CEL0}$\cite{RN47}} & \multicolumn{2}{c|}{$\rm LocNet$\cite{RN62}} & \multicolumn{2}{c|}{$\rm LocNet\ v2$} & \multicolumn{2}{c}{$\rm PiLocNet_g$} \\ \hline
Density & \multicolumn{1}{l}{Recall} & \multicolumn{1}{l|}{Precision} & Recall & \multicolumn{1}{l|}{Precision} & Recall & \multicolumn{1}{l|}{Precision} & Recall & Precision \\ \hline
10 & 95.80\% & 79.72\% & \textbf{96.00\%} & 92.15\% & 93.60\% & 92.60\% & 93.60\% & \textbf{92.62\%} \\
15 & 93.20\% & 77.68\% & \textbf{95.60\%} & 87.40\% & 93.73\% & 88.99\% & 94.27\% & \textbf{89.55\%} \\
20 & 89.30\% & 72.12\% & \textbf{92.95\%} & 81.63\% & 92.00\% & 85.21\% & 92.15\% & \textbf{85.84\%} \\
30 & 87.20\% & 58.77\% & 88.10\% & 72.56\% & 88.27\% & 79.15\% & \textbf{88.30\%} & \textbf{80.12\%} \\
40 & 77.40\% & 52.87\% & 84.28\% & 63.51\% & 85.12\% & 72.44\% & \textbf{85.23\%} & \textbf{73.06\%} \\
\hline
Average & 88.58\% & 68.23\% & \textbf{91.39\%} & 79.45\% & 90.54\% & 83.68\% & 90.71\% & \textbf{84.24\%} \\ \hline
\end{tabular}
}
\label{Gaussian main}
\end{table}

\begin{table}[ht]
\centering
\captionsetup{font={footnotesize,bf, sf,stretch=1},justification=centering}
\caption{Evaluation results of KL-NC, LocNet, LocNet v2, PiLocNet for RPSF images with Poisson noise  }
\resizebox{\textwidth}{!}{
\begin{tabular}{c|cc|cc|cc|cc}
\hline
 & \multicolumn{2}{c|}{$\rm KL-NC$\cite{RN47}} & \multicolumn{2}{c|}{$\rm LocNet$ \cite{RN62}} & \multicolumn{2}{c|}{$\rm LocNet\ v2$} & \multicolumn{2}{c}{$\rm PiLocNet_p$} \\ \hline
Density & \multicolumn{1}{l}{Recall} & \multicolumn{1}{l|}{Precision} & \multicolumn{1}{l}{Recall} & \multicolumn{1}{l|}{Precision} & Recall & Precision & Recall & Precision \\ \hline
10 & 99.20\% & 95.00\% & 98.90\% & 96.28\% & 99.20\% & 98.63\% & \textbf{99.30\%} & \textbf{99.17\%} \\
15 & 98.80\% & 89.18\% & 98.87\% & 95.54\% & 99.13\% & \textbf{98.99\%} & \textbf{99.33\%} & \textbf{98.99\%} \\
20 & 97.55\% & 85.02\% & 98.00\% & 94.45\% & 97.99\% & \textbf{98.70\%} & \textbf{98.95\%} & 97.85\% \\
30 & 97.30\% & 79.54\% & 96.87\% & 93.97\% & 97.53\% & 95.91\% & \textbf{97.80\%} & \textbf{96.67\%} \\
40 & 95.58\% & 73.64\% & 95.00\% & 90.59\% & 96.20\% & 93.83\% & \textbf{96.43\%} & \textbf{94.17\%} \\
\hline
Average & 97.69\% & 84.48\% & 97.53\% & 94.17\% & 98.15\% & 97.07\% & \textbf{98.36\%} & \textbf{97.37\%}\\
\hline
\end{tabular}}
\label{Poisson main}
\end{table}

The comprehensive outcomes are presented in Tables \ref{Gaussian main} and \ref{Poisson main}, delineating the performance metrics across the aforementioned methods under Gaussian and Poisson noise, respectively. The percentages shown in bold font in each row are the best ones that we obtained for recall and precision for the corresponding source density for the four methods. We restrict our attention here to only those images in which the number of point sources is either 10, 15, 20, 30, or 40 in order to have a more meaningful comparison with the already published results of the KL-NC \cite{RN47} and $\ell_2-{\rm CEL0}$ \cite{RN47} optimization approaches. 

The tabulated results show that both LocNet and PiLocNet, as neural network-based methods, substantially outperform the variational approach in handling images with either Gaussian or Poisson noise. Notably, PiLocNet, with its physics-informed design, shows typically the most impressive results, leading to the highest overall performance metrics for both noise cases. Specifically, PiLocNet improves precision by approximately 0.6\% over LocNet v2 in the Gaussian noise scenario and 0.3\% in the Poisson noise case. The recall rate has also improved, but not as much as the precision rate. The improvement in precision can be more substantial, however, at higher noise levels, as we will see later in Sec.\ref{VaryingNoise}.
This enhancement highlights the efficacy of incorporating physical knowledge into the neural network framework, particularly evident in the precision gains across both types of noise, affirming the value of physics-informed approaches in improving neural network predictions. 

It is noteworthy that LocNet v2 consistently demonstrates a certain improvement in precision rates compared to LocNet \cite{RN62}. This enhancement stems from LocNet v2's omission of the upsampling algorithm during its operation, which reduces the number of predicted point sources, particularly the false positives. This reduction of the false positives corresponds to a significant enhancement in the precision rates. 

\begin{figure}[htbp]
    \centering

    \begin{subfigure}[b]{0.24\textwidth}
        \includegraphics[width=1\textwidth]{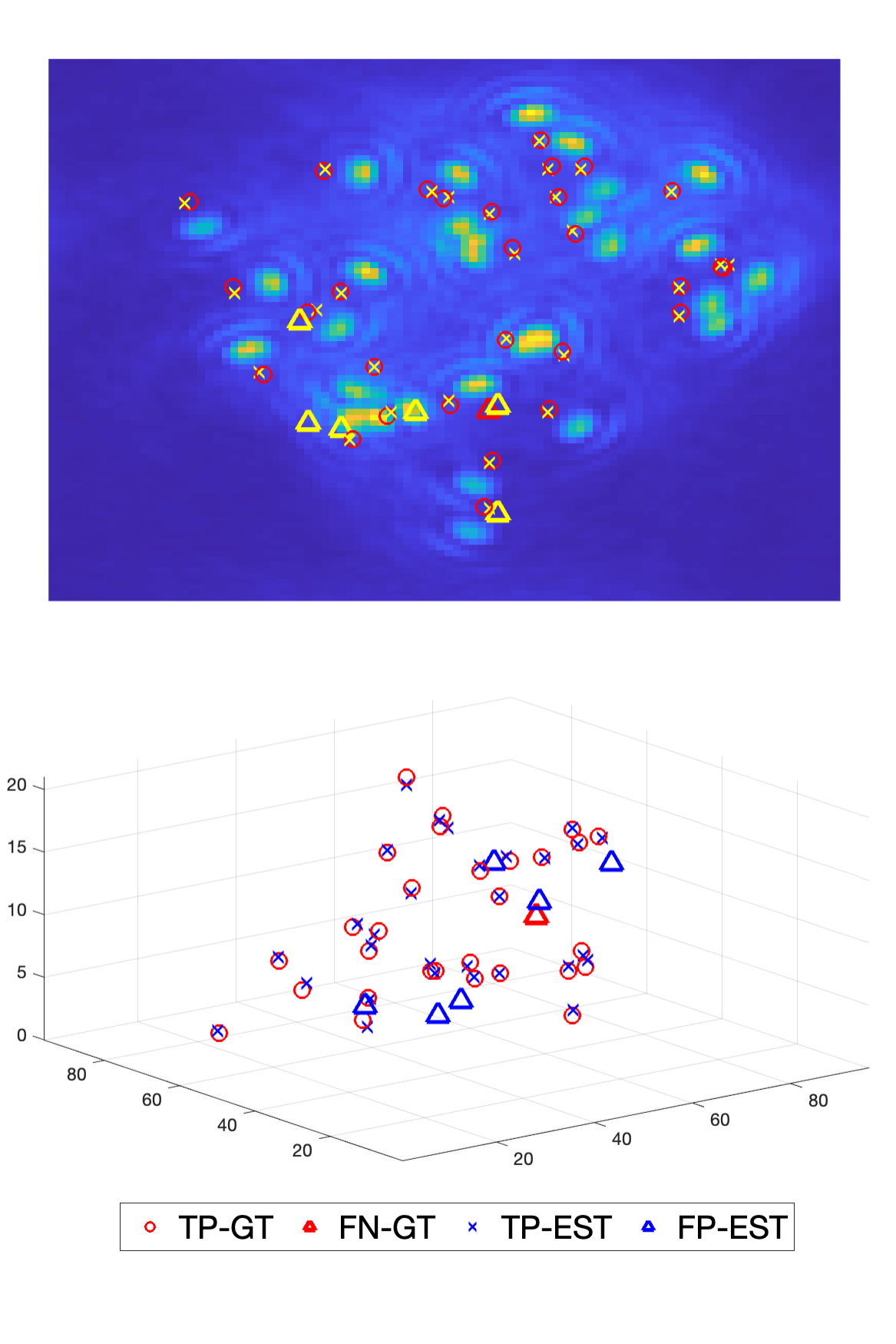}
        \caption{CEL0~\cite{RN47}}
        \label{cel0-2}
   \end{subfigure}
       \hspace{-0.05in}
    \begin{subfigure}[b]{0.24\textwidth}
         \includegraphics[width=1\textwidth]{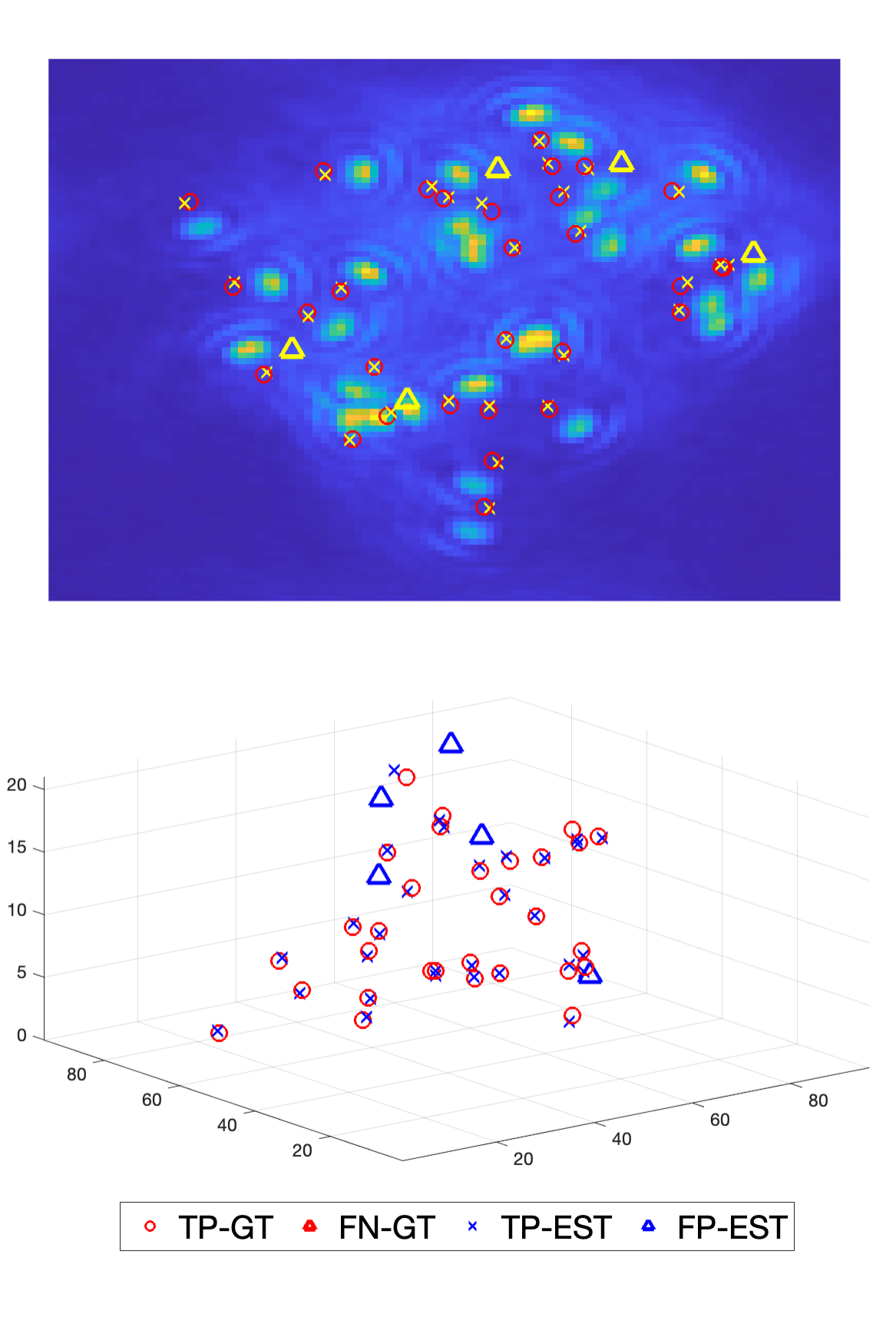}
        \caption{LocNet \cite{RN62} }
        \label{locnet-g-2}
    \end{subfigure}
    \hspace{-0.05in}
    \begin{subfigure}[b]{0.24\textwidth}
        \includegraphics[width=1\textwidth]{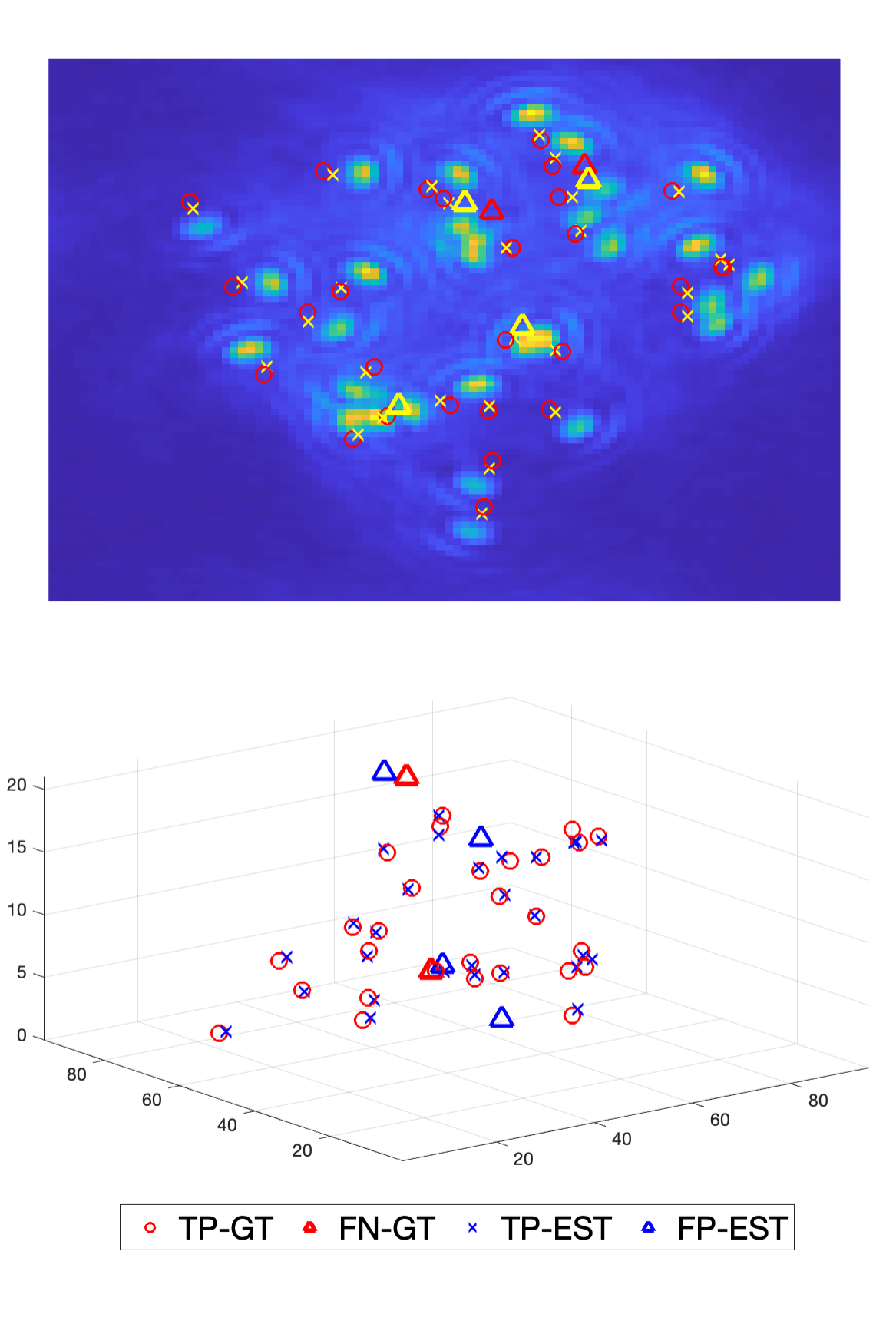}
        \caption{LocNet v2 }
        \label{locnet-g-v2-2}
    \end{subfigure}
    \hspace{-0.05in}
    \begin{subfigure}[b]{0.24\textwidth}
        \includegraphics[width=1\textwidth]{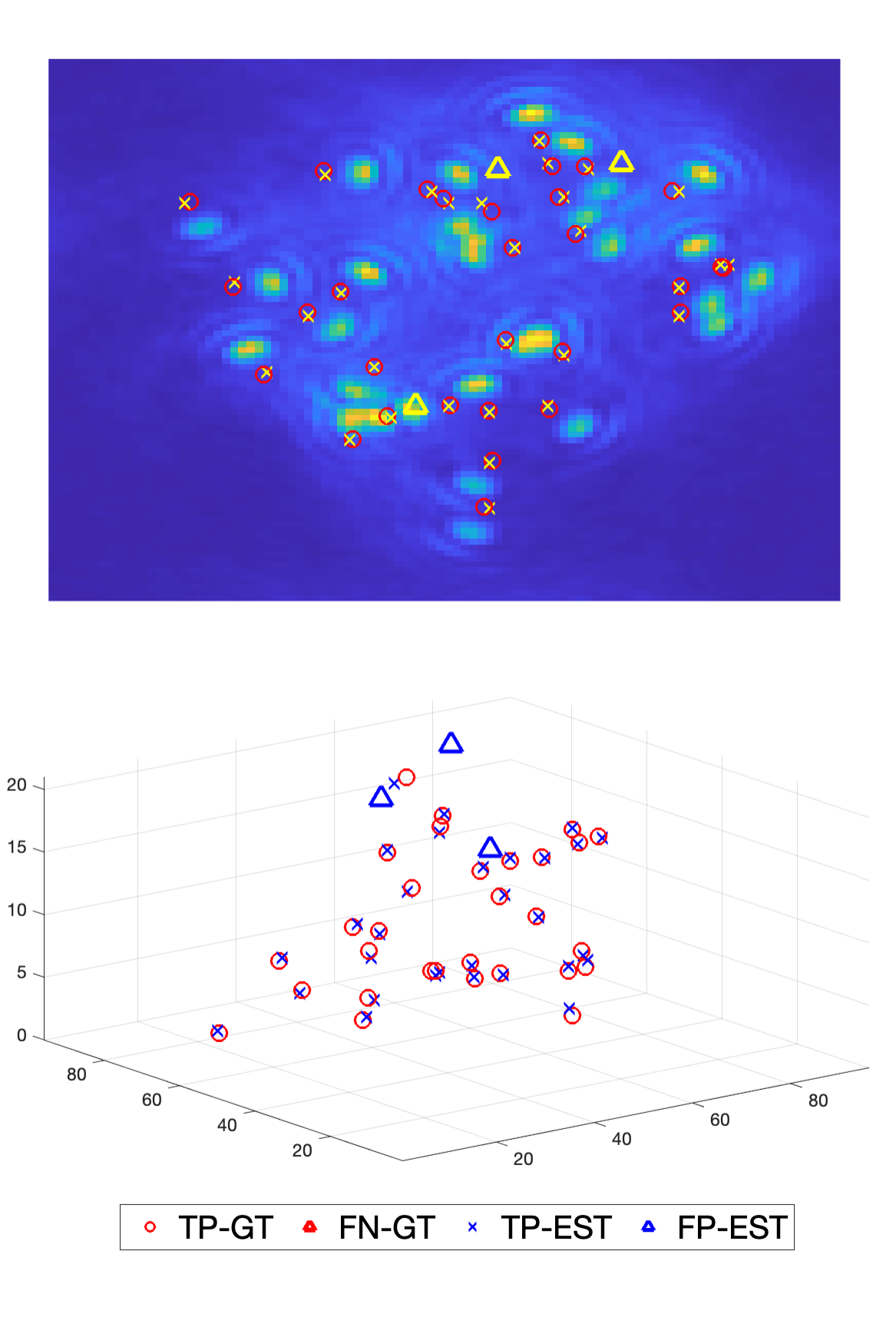}
        \caption{PiLocNet }
        \label{pilocnet-g-2}
    \end{subfigure}
    \caption{2D snapshot images (top) and 3D locations (bottom) for the 30-point-source case with Gaussian noise. The triangles denotes the missed matches. 
    }
    \label{4models_g}
\end{figure}

\begin{figure}[htbp]
    \centering

    \begin{subfigure}[b]{0.24\textwidth}
        \includegraphics[width=1\textwidth]{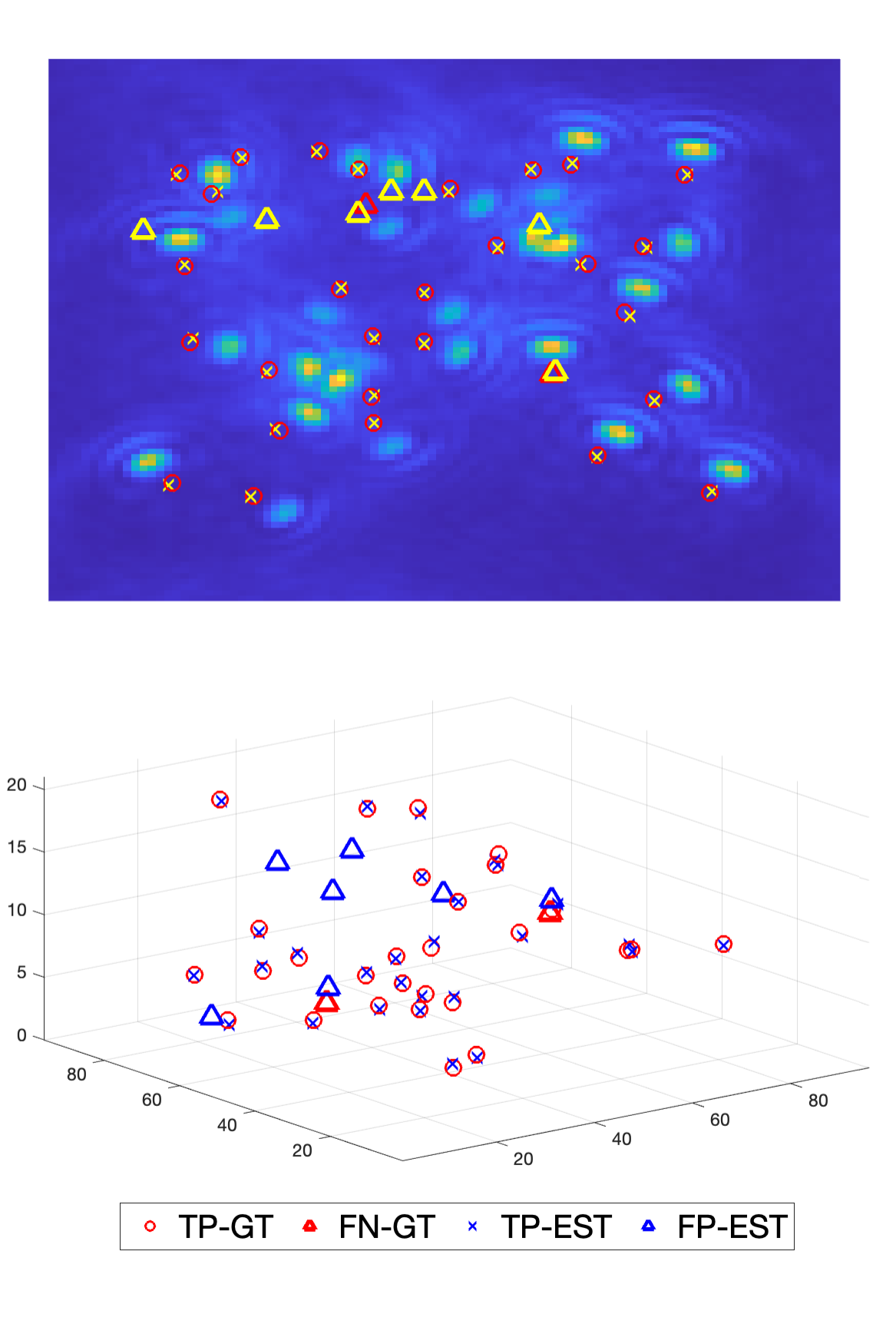}
        \caption{KL-NC~\cite{RN47}}
        \label{kl-nc-2}
   \end{subfigure}
       \hspace{-0.05in}
    \begin{subfigure}[b]{0.24\textwidth}
         \includegraphics[width=1\textwidth]{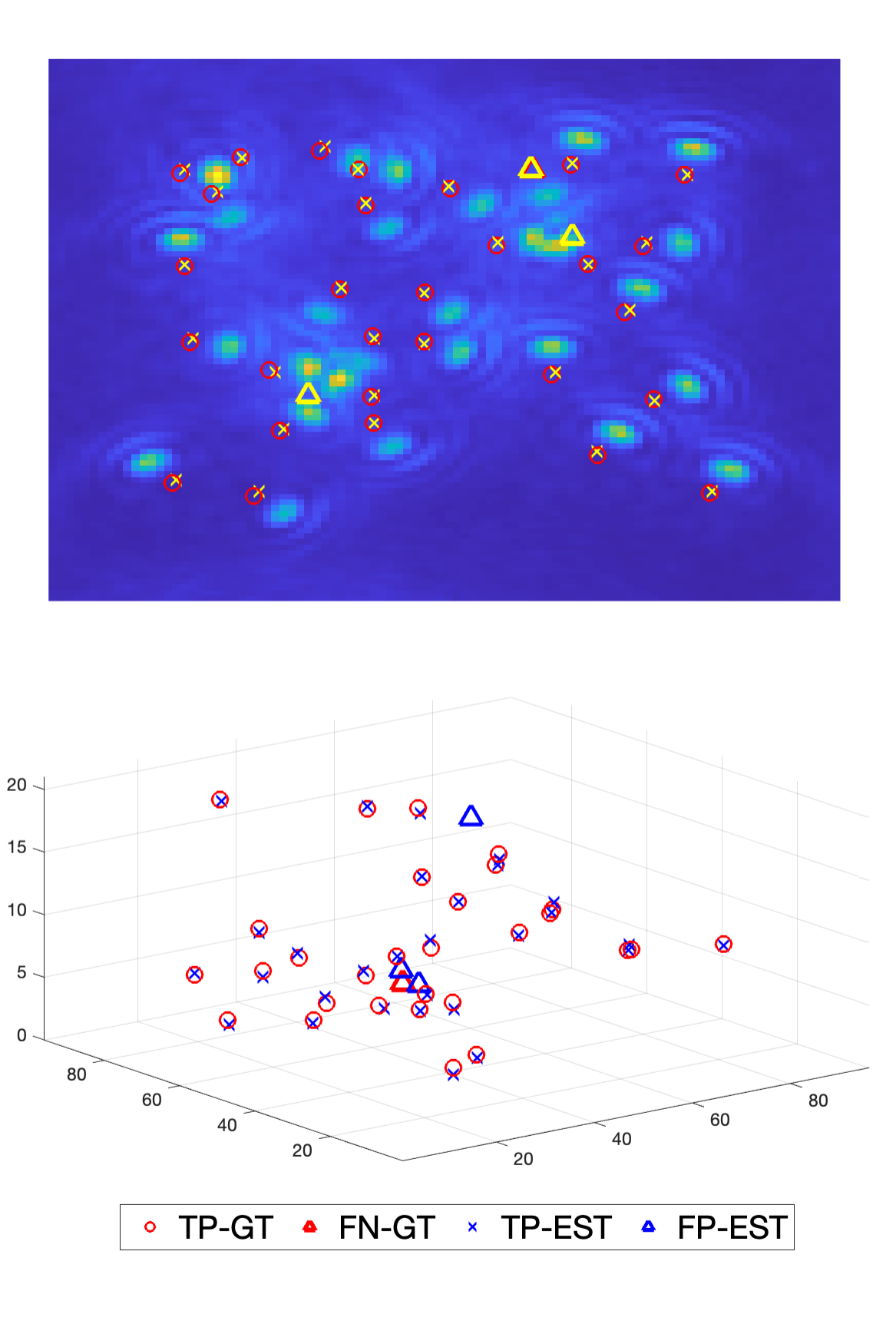}
        \caption{LocNet \cite{RN62}}
        \label{locnet-2}
    \end{subfigure}
    \hspace{-0.05in}
    \begin{subfigure}[b]{0.24\textwidth}
        \includegraphics[width=1\textwidth]{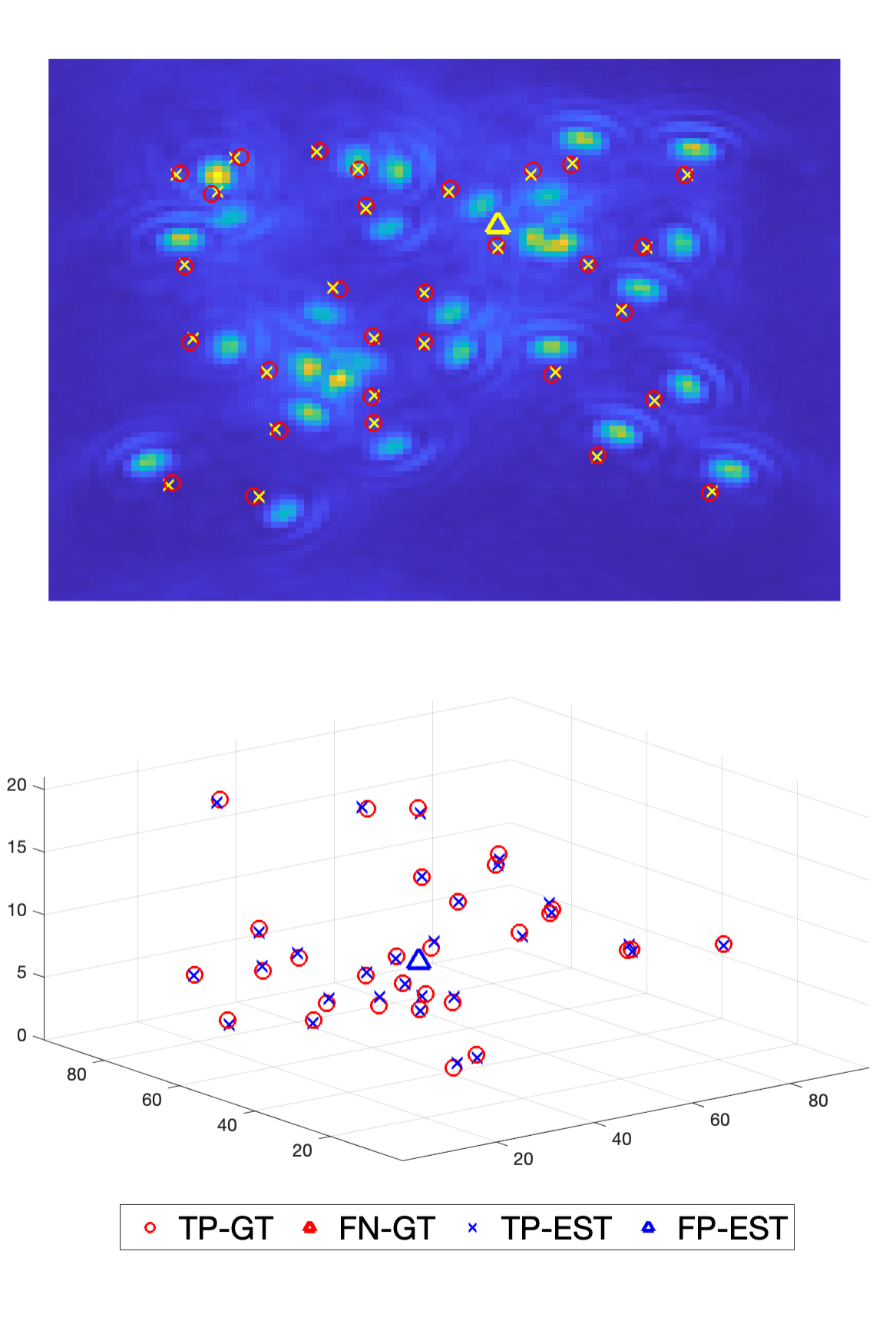}
        \caption{LocNet v2 }
        \label{locnetv2-2}
    \end{subfigure}
    \hspace{-0.05in}
    \begin{subfigure}[b]{0.24\textwidth}
        \includegraphics[width=1\textwidth]{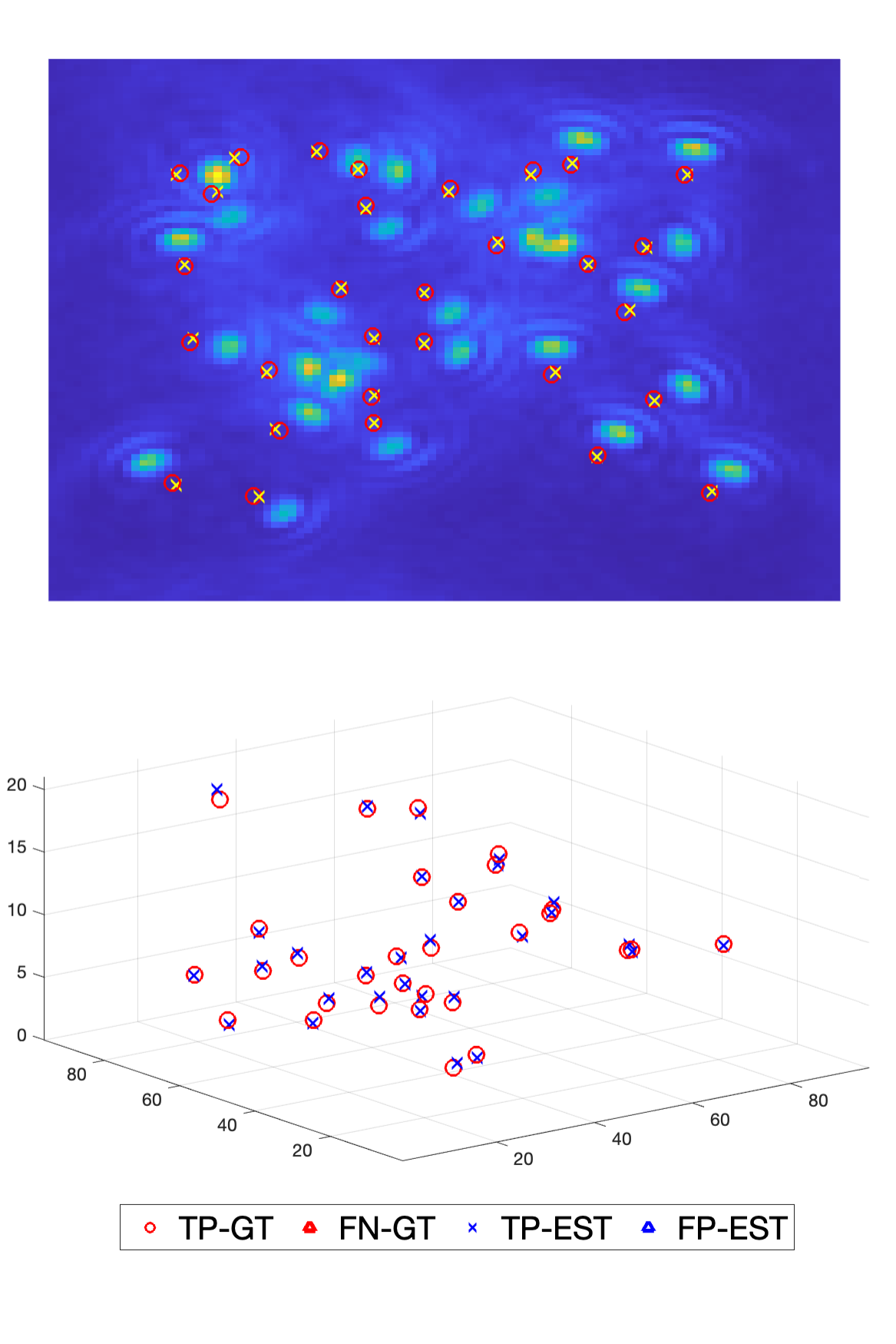}
        \caption{PiLocNet }
        \label{pilocnet-2}
    \end{subfigure}
    \caption{2D snapshot images (top) and 3D locations (bottom) for the 30-point-source case with Poisson noise. The triangles denote the missed matches (see text for more details).}
    \label{4models_p}
\end{figure}

Gaussian and Poisson noise cases have been shown in Fig.~\ref{4models_g} and Fig.~\ref{4models_p}. The first row in each figure refers to the same specific 2D snapshot where ``o'' labels the $(x,y)$ positions of the ground-truth point sources, ``x'' labels the estimated point sources according to the method used, and ``$\triangle$'' represents a mismatch, with the red and yellow colors labeling false-negative and false-positive sources, respectively. The second row shows the locations in 3D grids where the ground-truth point sources are in red markers with red ``$\triangle$'' being false-negative and red ``o'' being true-positive. The estimated source positions are in yellow for the 2D snapshots in the first row and in blue for the 3D grids in the second row, with ``$\triangle$'' denoting false-positive and ``x'' denoting true-positive. 
It is evident that compared to the original LocNet, LocNet v2 exhibits lower prediction errors. However, it fails at times to predict a ground-truth point source. The fact that when two or more ground-truth point sources are located closely, LocNet v2, having abandoned the upsampling process, is less sensitive to locating such densely packed point sources is the root of such failures. By contrast, since the loss function of PiLocNet incorporates additional information, it more effectively mitigates these errors.

\subsection{Contributions of the data-fitting and regularization terms}

Note that the physics-informed loss function of PiLocNet, is represented by Eq.~\ref{eq:main_loss}, 
$\mathcal{L}_{\rm {PiLocNet}}=w_1\mathcal{D}+\ w_2\mathcal{R}+w_3 \rm MSE$,
where $\mathcal{D}$ and $\mathcal{R}$ are model specific terms.
Setting the first two weights to zero, $w_1 = 0, w_2 = 0$,  reduces PiLocNet to LocNet v2. For this section, we set up control groups to study the individual contribution of each added term within the loss function, presenting our results in \Cref{contribution}. 

 For the Gaussian case, adding $\mathcal{D}$ to MSE improves the average recall rate from 90.54\% to 90.67\%, but the average precision drops from 83.68\% to 82.38\%. When only $\mathcal{R}$ is added to MSE, we increase the precision from 83.68\% to 84.19\%, while the average recall is not as good as that of the group in which only $\mathcal{D}$ has been added to MSE. For PiLocNet, we find the best average recall and precision rates. 
Similarly, for the Poisson case, adding $\mathcal{D}$ to MSE improves average recall from 98.15\% to 98.27\%, the latter being the best recall result among all groups. However, its precision decreases from 97.07\% to 96.33\%. Combining $\mathcal{D}$, $\mathcal{R}$, and MSE, we once again achieve the best average recall and precision rates. 

\begin{figure}[htbp]
    \centering
    \begin{subfigure}[b]{0.23\textwidth}
        \includegraphics[width=1\textwidth]{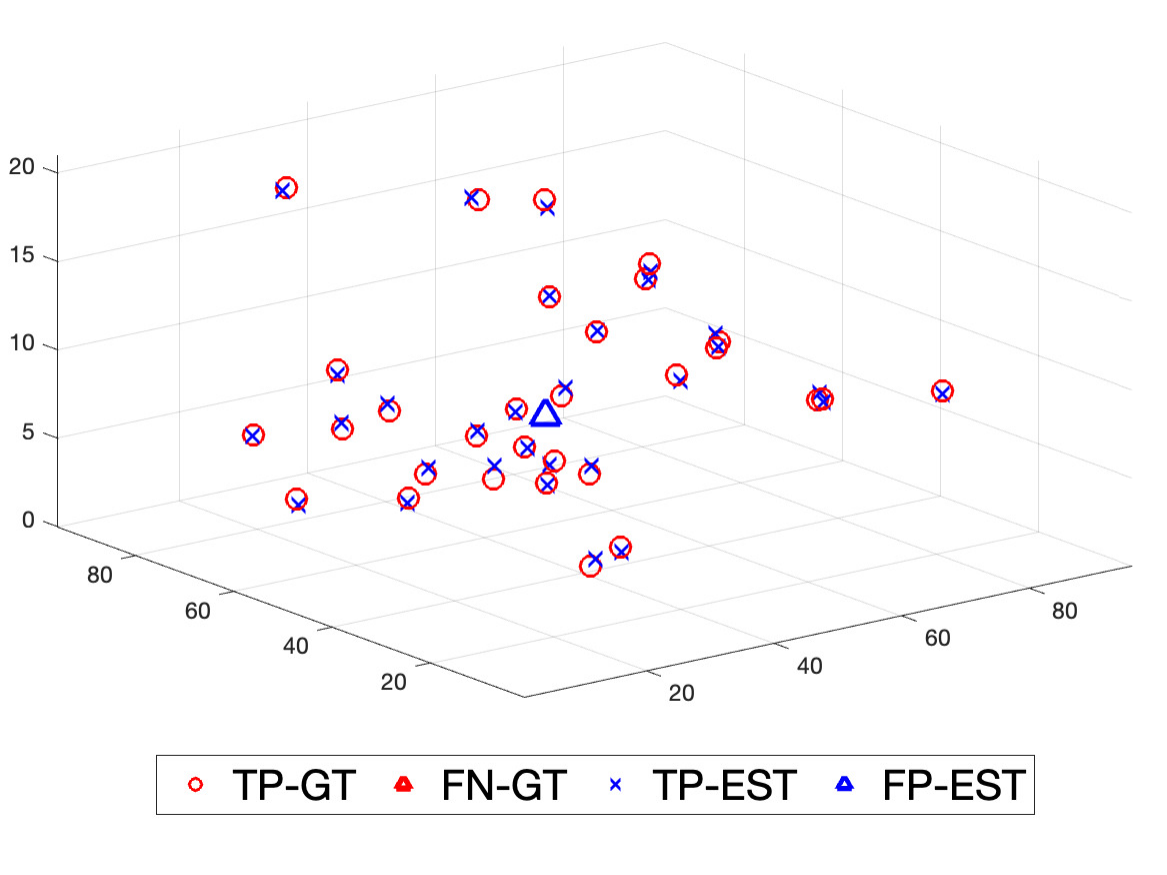}
        \caption{MSE only.}
        \label{0-0-1}
    \end{subfigure}
    \hspace{0in}
    \begin{subfigure}[b]{0.23\textwidth}
         \includegraphics[width=1\textwidth]{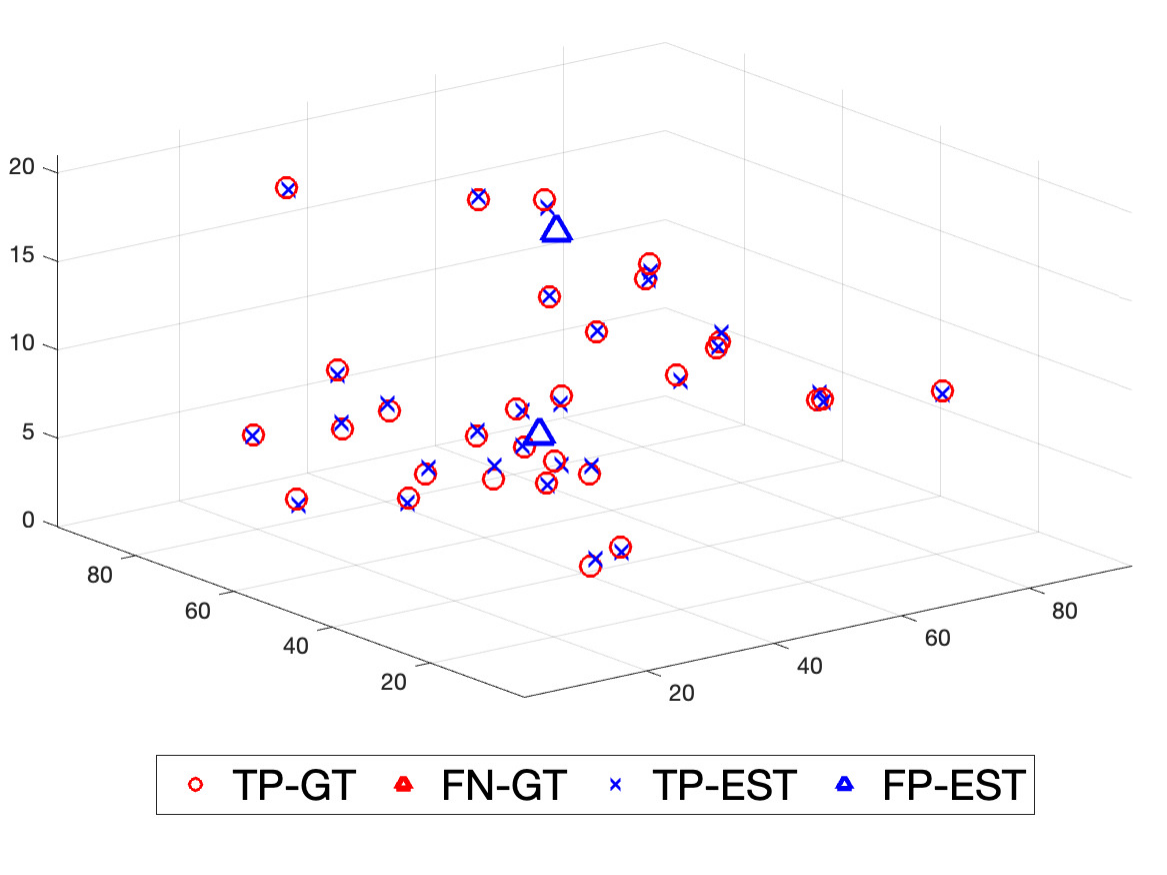}
        \caption{MSE + $\mathcal{D}$.}
        \label{1-0-500}
    \end{subfigure}
    \hspace{0in}
    \begin{subfigure}[b]{0.23\textwidth}
        \includegraphics[width=1\textwidth]{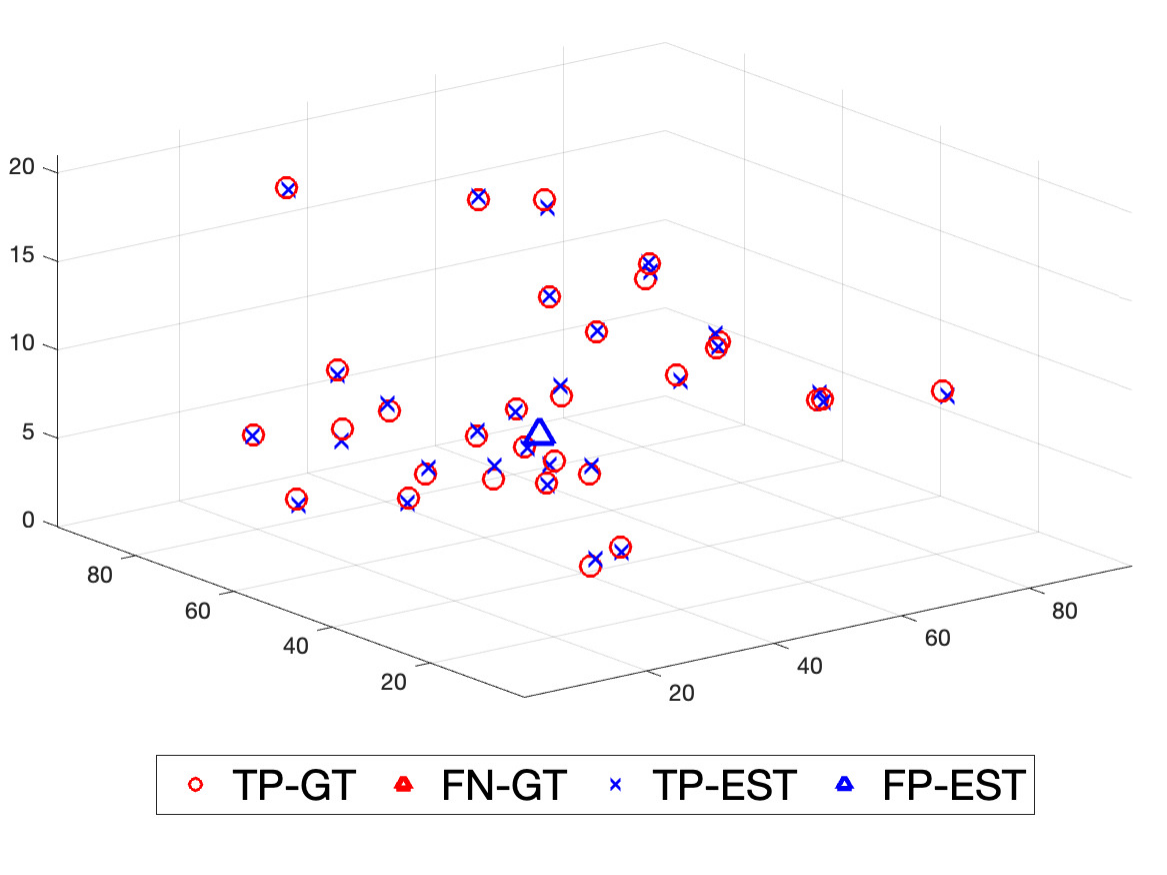}
        \caption{MSE + $\mathcal{R}$.}
        \label{0-1-500}
    \end{subfigure}
    \hspace{0in}
    \begin{subfigure}[b]{0.23\textwidth}
        \includegraphics[width=1\textwidth]{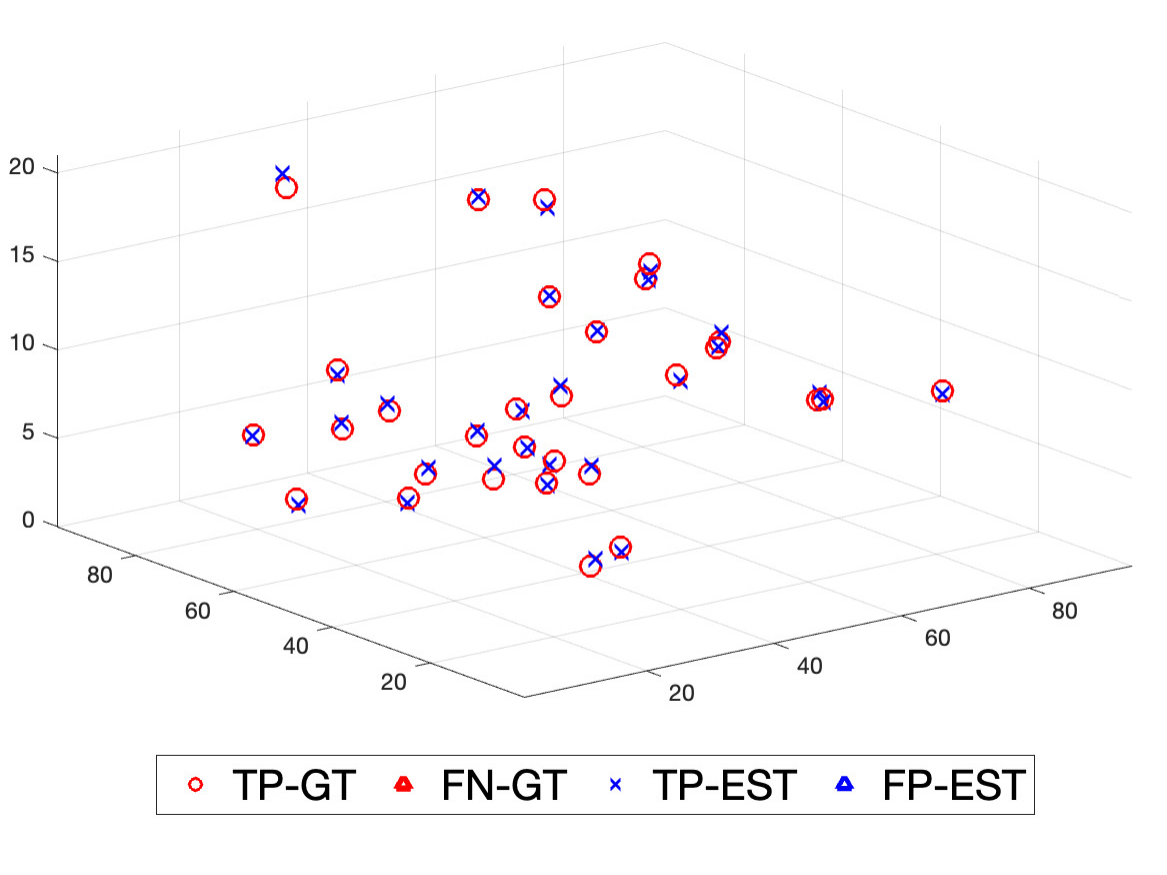}
        \caption{MSE + $\mathcal{D} + \mathcal{R}$.}
        \label{1-1-500}
    \end{subfigure}
    \medskip
    \caption{The effects of the different components in the loss function Eq.~\ref{eq:main_loss}. 
     The triangles denotes the missed matches.
    }
    \label{comparing methods}
\end{figure}


The effect of each added term in the loss function can be visualized by the restored 3D tensor in Fig.~\ref{comparing methods}. 
Simply combining MSE and $\mathcal{D}$ turns some false negatives into true positives, while also introducing some false positives, as evidenced by Fig.~\ref{1-0-500}, which displays a higher count of falsely estimated points compared to Fig.~\ref{0-0-1}. In contrast, the regularization term $\mathcal{R}$ tends to elevate the precision rate, as its inclusion within the variational framework acts to control sparsity, thereby mitigating the occurrence of undesired false positives. Fig.~\ref{0-1-500} is a compelling illustration of how the incorporation of regularized terms alone can diminish the occurrence of falsely estimated points. By leveraging both components, PiLocNet balances these effects, leading to an overall enhancement, as seen in Fig.~\ref{1-1-500}.

\begin{table}[htbp]
\centering
\footnotesize
\captionsetup{font={footnotesize,bf, sf,stretch=1},justification=centering}
  \caption{Effectiveness of data-fitting and regularization terms for Gaussian and Poisson noised images. The average values of precision and recall among different density cases are shown. }
\begin{tabular}{ccc|cc|cc}
\hline
\multicolumn{3}{c|}{\textbf{Components}} & \multicolumn{2}{c|}{\textbf{Gaussian noise}} & \multicolumn{2}{c}{\textbf{Poisson noise}} \\ \hline
$\mathcal{D}$ & $\mathcal{R}$ & {\rm MSE} & Recall & Precision & Recall & Precision \\ \hline
\texttimes & \texttimes & \checkmark & 90.54\% & 83.68\% & 98.15\% & 97.07\% \\
\checkmark & \texttimes & \checkmark & 90.67\% & 82.38\% & 98.27\% & 96.33\% \\
\texttimes & \checkmark & \checkmark & 89.78\% & 84.19\% & 97.96\% & 96.21\% \\
\checkmark & \checkmark & \checkmark & \textbf{90.71\%} & \textbf{84.24\%} & \textbf{98.36\%} & \textbf{97.37\%}\\
\hline
\end{tabular}
\label{contribution}
\end{table}


\subsection{Robustness on Noise Level}
\label{VaryingNoise}



\begin{figure}[htbp]
    \centering
    \begin{subfigure}[b]{0.45\textwidth}
        \includegraphics[width=1\textwidth]{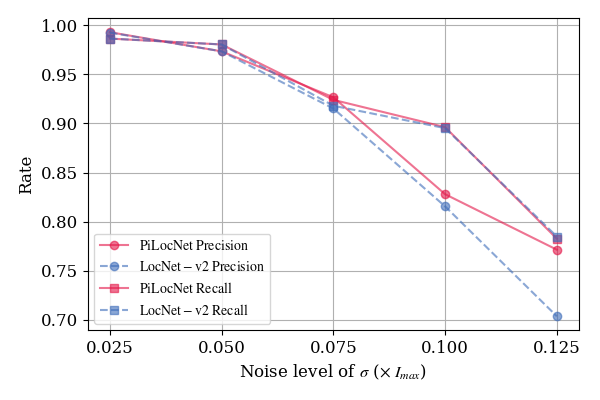}
        \caption{Gaussian}
        \label{pilocnet_gaussian_p25}
    \end{subfigure}
    \begin{subfigure}[b]{0.45\textwidth}
        \includegraphics[width=1\textwidth]{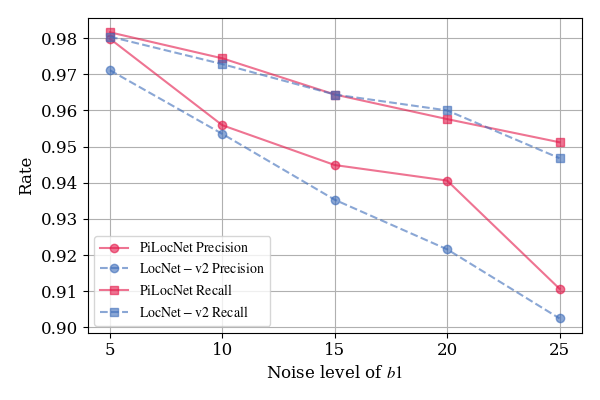}
        \caption{Poisson}
        \label{pilocnet_poisson_p25}
    \end{subfigure}
    \medskip
    
    \caption{Precision and recall rates for 25 point sources at 5 different noise levels for the Gaussian and Poisson noise cases. 
    }
    \label{robustness2}
\end{figure}

We also conducted a comprehensive assessment of the robustness of PiLocNet's performance, relative to LocNet v2's, across varying noise levels for the two models of noise considered here. 
We chose the number of point sources to be 25 for this purpose. In the case of Gaussian noise, we evaluated performance across five different noise levels, as depicted in Fig.~\ref{pilocnet_gaussian_p25}, where the horizontal axis represents the noise level, with the noise standard deviation, $\sigma$, being the indicated value multiplied by $I_{\rm max}$ and the background $b$ fixed at $b = 5$ for each noise level. The results involving Poisson noise are illustrated in Fig.~\ref{pilocnet_poisson_p25}, where the horizontal axis represents the background photon count, $b$. Across both noise types, PiLocNet consistently outperformed LocNet v2, demonstrating superior precision while maintaining a very similar recall rate (nearly overlapping red and blue $\Box$'s). Notably, as the noise level increased, PiLocNet exhibited even more significant precision gains over LocNet v2. For instance, under Gaussian noise with a variance of $\sigma = 0.100 \times I_{\rm max}$, PiLocNet's precision was higher by about 1\%, while at $\sigma = 0.125 \times I_{\rm max}$, its precision gain exceeded 6\%. 

\section{Conclusions}\label{sec:conclusion}


The new method, PiLocNet, that we have proposed here adds useful physical information to the neural network by adding to the conventional network loss function data-fitting and regularization terms that match the noise model governing the observed image data. As we have shown, this greatly improves the network performance. The principal change in the network architecture from LocNet, namely the removal of upsampling and a coarsening of the 3D grid, leads to a reduction of false positives while greatly shortening the network training speed without sacrificing overall performance. PiLocNet outperforms previous methods, as we have demonstrated through robust validation processes. 

In the modified loss function of PiLocNet, the data-fitting term $\mathcal{D}$ containing the additional PSF matrix information tends to recover point predictions that would otherwise be missed. The regularization term $\mathcal{R}$, on the other hand, exploits sparsity to reduce the occurrence of false positives. These two effects improve the overall network performance by reducing the rates of both false negatives and false positives. 

Neural networks, when well trained, excel at making predictions from highly complex datasets, while variational methods critically employ physical information about the PSF and noise model as well as regularization to avoid overfitting. PiLocNet combines the strengths of both these approaches. By embedding the forward model directly into a neural network, PiLocNet can implement a broad range of PSFs and imaging challenges when the forward model is accurately known. In our future work, we will explore improved data-fitting methodologies to further enhance the performance of PiLocNet. Additionally, we plan to extend the application of our methodology beyond simulated experiments to real-world datasets.

\begin{backmatter}
\bmsection{Funding}
HKRGC Grants Nos.
N\_CityU214/19, CityU11301120, C1013-21GF, and CityU11309922; 
the Natural Science Foundation of China No. 12201286; 
Guangdong Basic and Applied Research Foundation 2024A1515012347;
Shenzhen Science and Technology Program 20231115165836001;  CityU Grant 9380101.
\bmsection{Disclosures}
The authors declare no conflicts of interest.
\bmsection{Data Availability}
Data underlying the results presented in this paper are not publicly available at this time but may be obtained from the authors upon reasonable request.
\end{backmatter}

\bibliography{main_ao}
\end{document}